\pdfoutput=1

\documentclass[11pt]{article}

\usepackage[final]{acl}

\usepackage{times}
\usepackage{latexsym}

\usepackage[T1]{fontenc}

\usepackage[utf8]{inputenc}

\usepackage{microtype}
\usepackage{amsmath}
\usepackage{amssymb}
\usepackage{graphicx}
\usepackage{bm}
\usepackage{diagbox}
\usepackage{xcolor}
\usepackage{inconsolata}

\usepackage{booktabs} 
\usepackage{multirow}

\title{Decoding the Echoes of Vision from fMRI: Memory Disentangling for Past Semantic Information} 

\author{Runze Xia, Congchi Yin, Piji Li$^{\ast}$\\ 
\textsuperscript{\rm 1} College of Computer Science and Technology,\\
Nanjing University of Aeronautics and Astronautics, China\\
\textsuperscript{\rm 2} MIIT Key Laboratory of Pattern Analysis and Machine Intelligence, Nanjing, China\\
\texttt{\{xiarunze,congchiyin,pjli\}@nuaa.edu.cn}}

\begin{document}
\maketitle
\renewcommand{\thefootnote}{\fnsymbol{footnote}}
\footnotetext[1]{Corresponding author.}
\renewcommand{\thefootnote}{\arabic{footnote}}

\begin{abstract}
The human visual system is capable of processing continuous streams of visual information, but how the brain encodes and retrieves recent visual memories during continuous visual processing remains unexplored. This study investigates the capacity of working memory to retain past information under continuous visual stimuli. And then we propose a new task \textbf{Memory Disentangling}, which aims to extract and decode past information from fMRI signals. To address the issue of interference from past memory information, we design a disentangled contrastive learning method inspired by the phenomenon of proactive interference. This method separates the information between adjacent fMRI signals into current and past components and decodes them into image descriptions. Experimental results demonstrate that this method effectively disentangles the information within fMRI signals. This research could advance brain-computer interfaces and mitigate the problem of low temporal resolution in fMRI. \footnote{Code is available at \url{https://github.com/xiaRunZe/Memory-Disentangling}}

\end{abstract}

\section{Introduction}

The human visual system is highly intricate and plays a foundamental role in daily lives \cite{loomis2018sensory}. Exploring and comprehending this system is a key objective for researchers in the fields of neuroscience and artificial intelligence \cite{clark2013whatever,herreras2010cognitive}. One particularly intriguing question pertains how the brain processes and retrieves recent visual memories, which holds  significant implications for Brain-Computer Interfaces (BCIs) and cognitive neuroscience \cite{logothetis2008we,ranganath2005directing,marr2010vision}.

In recent years, functional magnetic resonance imaging (fMRI), a revolutionary non-invasive neuroimaging technique \cite{ogawa1990brain}, has become indispensable for studying brain function and cognitive processes by detecting blood flow changes associated with neural activity through blood-oxygen-level-dependent (BOLD) contrast \cite{bandettini1992time}. With its high spatial resolution, fMRI rapidly advances neuroscience researches \cite{glover2011overview}. Therefore, fMRI provides our research with a unique perspective to explore the relationship between brain activity and memory functions. 
\begin{figure}[t]
    \centering
    \begin{minipage}[t]{\textwidth}
 \includegraphics[scale=0.5]{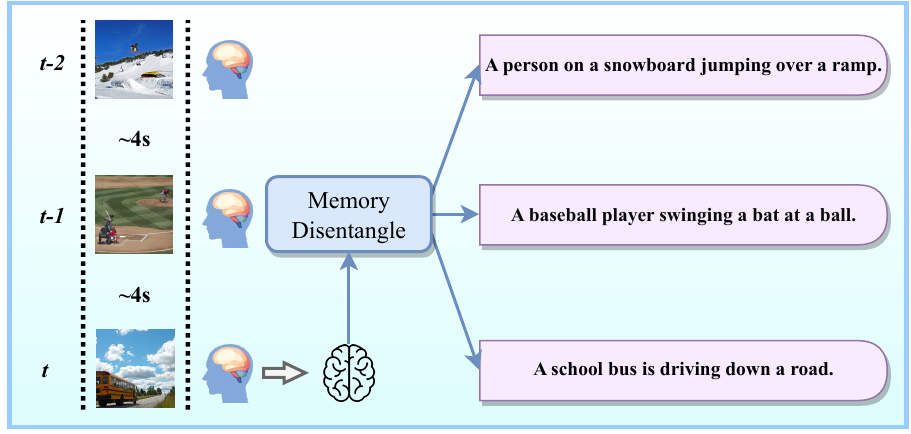}
    \end{minipage}
    \caption{The schematic diagram of Memory Disentangling based on decoding image semantic information.}  
    \vspace{-1em}
    \label{fig:intro}
\end{figure}

While significant progress \cite{xia2024dream, ozcelik2022reconstruction, takagi2023high} has been made in fMRI studies involving static visual stimuli, research on continuous visual stimuli remains largely unexplored. In real-world scenarios, visual experiences are rarely isolated and static. Instead, our brain continuously processes streams of visual information, necessitating the tracking of scene changes and retention of critical visual details to support decision-making \cite{yin2020direct}. Although studies under static visual stimuli have provided valuable insights into the visual system \cite{rossiter2001brain}, they neglect the continuity and dynamics of visual information, limiting our understanding of how the brain encodes memory within a continuous visual flow. Hence, it is crucial to explore how the brain processes memory information and how the representation of memory changes within the brain under continuous visual stimuli. Therefore, we endeavor to analyze memory under continuous visual stimuli to advance research on how the brain processes continuous visual stimuli. 


Memory is a core component of human cognitive architecture, allowing us to store and recall past experiences. In visual perception, memory involves not only encoding individual scenes but also integrating and updating continuous streams of visual information \cite{miller1956magical,cowan2001magical}. According to working memory theory, the human brain can temporarily store and manipulate information \cite{baddeley1992working}. However, the capacity of short-term memory especially working memory is limited, typically around a few items \cite{luck1997capacity}.

One of the motivations of our study is to explore the ability of working memory to retain past information under continuous visual stimuli by analyzing fMRI signals, particularly the semantic information of images from the past few moments. We use two data analysis techniques, ridge regression analysis and trial-wise representational similarity analysis (RSA), to assess the correlation between fMRI signals and visual stimuli from different past time points. We find that the correlation between fMRI signals and past semantic information gradually decreases over time, retaining at most 3-4 items, which aligns with the characteristics of working memory \cite{luck1997capacity, baddeley1992working}.

Based on the analysis of the correlation between fMRI signals and past visual stimuli, we propose the new \textbf{Memory Disentangling} task. This task aims to extract past visual stimuli information from brain activity and separate it from ongoing brain activity to mitigate the effects of proactive interference. To simplify this task, we focus on Memory Disentangling based on decoding the semantic information of these stimulus. Specifically, we decode the fMRI signals to get the semantic information of images both current and past moments. A schematic illustration of this process is shown in Figure \ref{fig:intro}. This task can contribute to advancing brain memory decoding research. Besides, by decoding past information, it also addresses the low temporal resolution issue inherent in fMRI signals.

To achieve this goal, we first propose a straightforward method for Memory Disentangling by employing multiple separate Multilayer Perceptron to map fMRI signals to semantic features of images at multiple time points. Additionally, inspired by proactive interference in working memory \cite{oberauer2001beyond, keppel1962proactive}, we introduce contrastive learning for disentangling. This method leverages relationships between consecutive pairs of fMRI signals to enhance the accuracy of extracting past information. Subsequently, we discuss how to transform these semantic features into intuitive, textual representations of semantic content and evaluate their effectiveness.

Our contributions are as follows:
\begin{itemize}
    \setlength{\itemsep}{0pt}
    \setlength{\parsep}{0pt}
    \setlength{\parskip}{0pt}
\item We analyze the capacity of working memory using fMRI signals and proposed a new task``Memory Disentangling'' based on these findings, aiming to decode past information from current brain signals and mitigate memory interference.
\item We introduce a memory disentangled contrastive learning method to accomplish the Memory Disentangling task, leveraging the theory of proactive interference to disentangle past memory information from current fMRI signals.
\item We conduct extensive experiments to validate the role of disentangled contrastive learning, demonstrating its effectiveness for mitigate memory interference and providing insights that guide future brain decoding tasks to consider the impact of past memories.
\vspace{-1em}
\end{itemize}

\begin{figure*}[t]
  \includegraphics[scale=0.92]{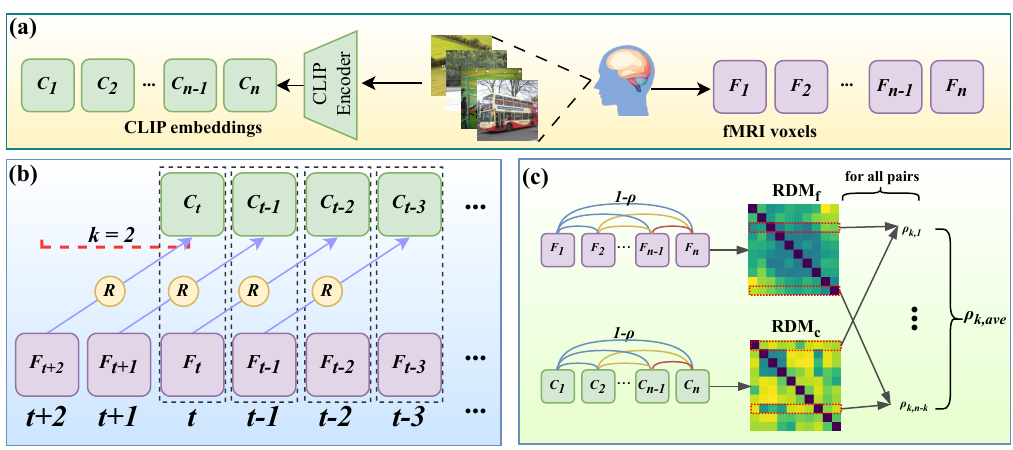} 
  \caption {Overview of visual memory analysis. (a) Acquisition of continuous visual stimuli data, including image embeddings and fMRI signals. (b) Ridge regression analysis for visual memory, where``R'' represents the ridge regression model, and $k$ is offset. The figure illustrates an example for $k=2$. (c) Trail-wise RSA, with the meaning of k remaining consistent with the previous context. Note that, for explanatory purposes, the size of the RDMs in the figure is illustrative and not representative of the actual size.}
  \vspace{-2em}
    \label{fig:ana}
\end{figure*}

\section{Analysis and Task Definition}
As previously mentioned, working memory has the ability to temporarily store and manipulate information. We investigate the capacity of working memory to retain past information based on fMRI signals under continuous visual stimuli. Based on this investigation, we propose a new task in the field of brain decoding, termed ``Memory Disentangling'', which will be detailed in Section \ref{sec:task}.
\begin{figure}[t]
  \includegraphics[width=\columnwidth]{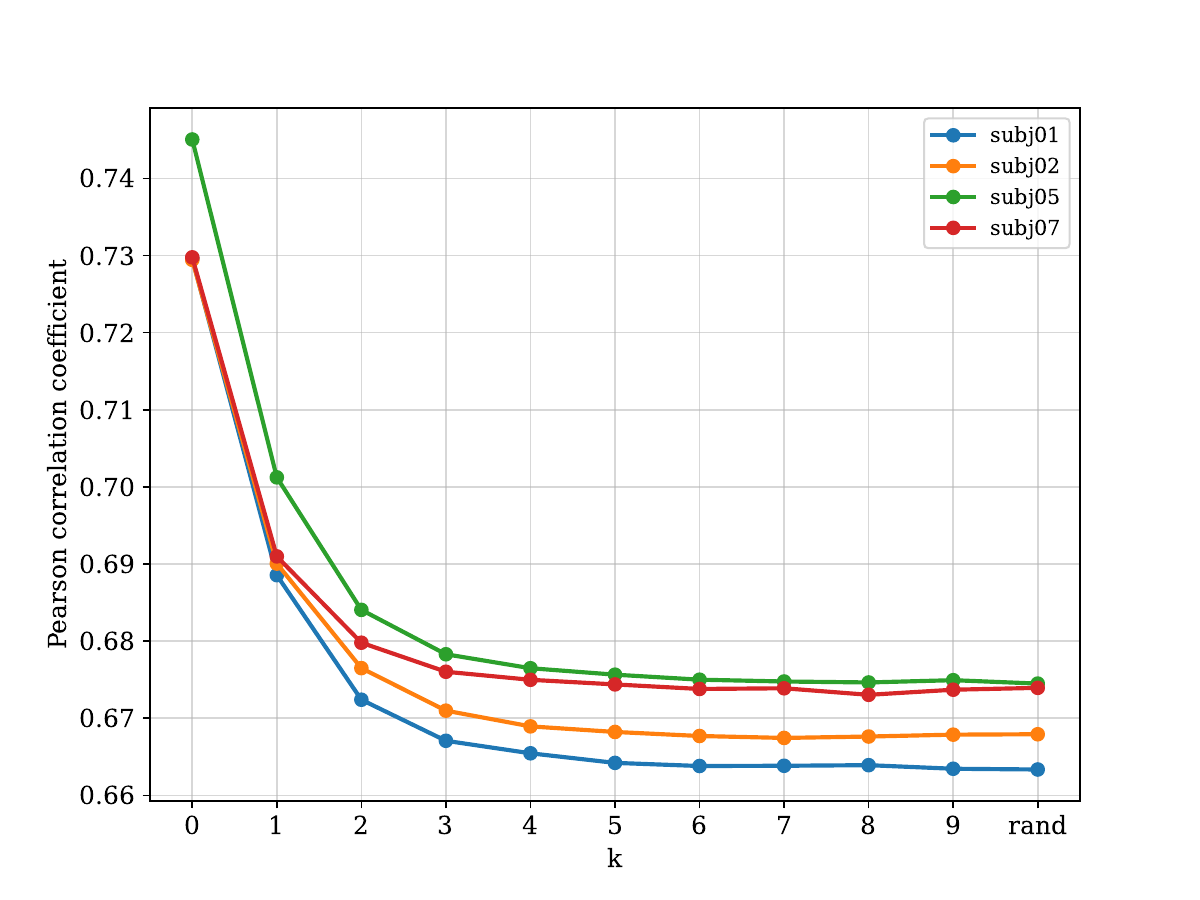} 
  \caption {The results of ridge regression analysis for four participants.}
\vspace{-1em}
    \label{fig:pcc}
\end{figure}
\begin{figure}[t]
  \includegraphics[width=\columnwidth]{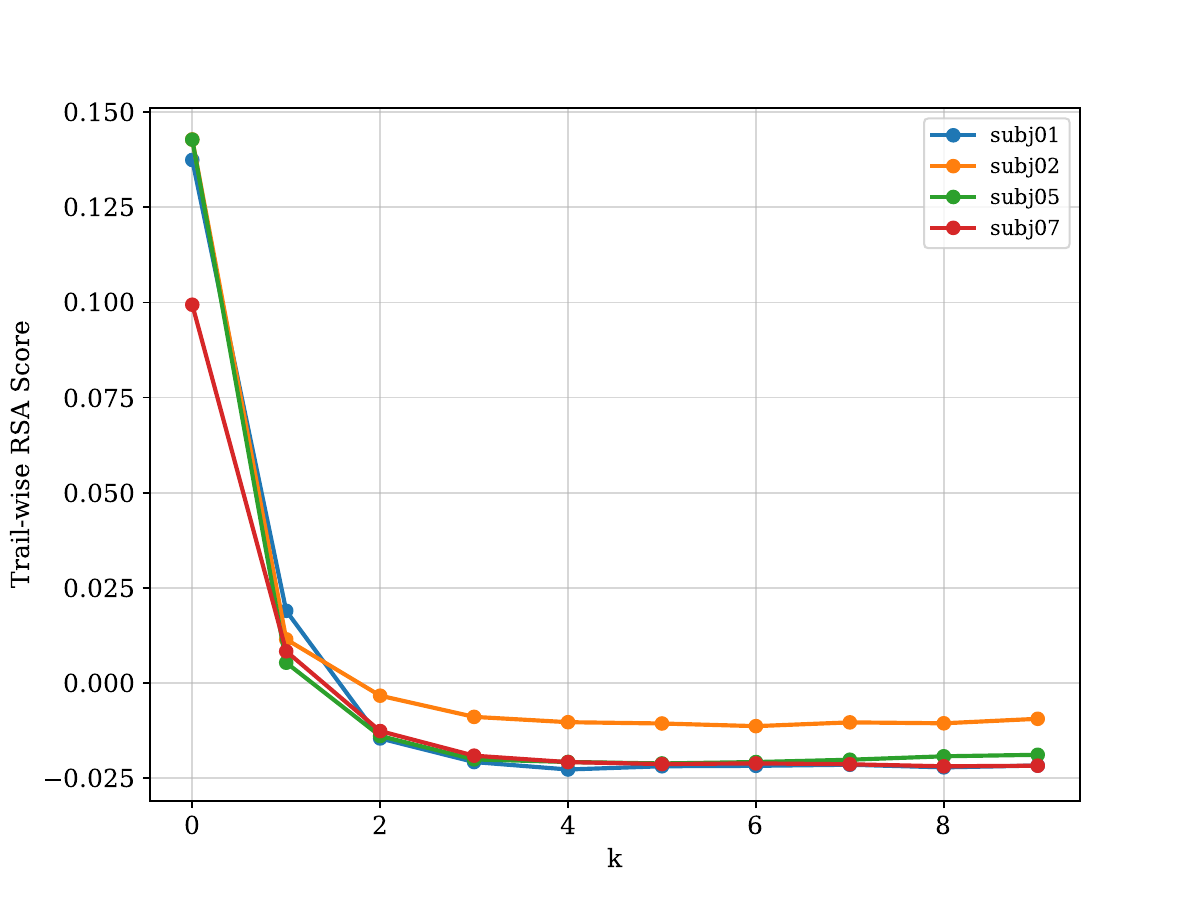} 
  \caption {The trail-wise RSA results for four participants.}
    \label{fig:rsa}
    \vspace{-1em}
\end{figure}
\vspace{-1em}
\subsection{Visual Memory Analysis} 
In this section, we explore whether fMRI signals can reflect past visual stimuli and the duration for which this information be retained in the fMRI signals. To achieve this goal, we employ ridge regression analysis and trial-wise RSA to examine the correlation between fMRI signals and visual stimuli from different past time points. The overview of this section is displayed in Figure \ref{fig:ana}.

\paragraph{Data Acquisition}
The Natural Scenes Dataset (NSD)~\cite{allen2022massive}, which is the largest fMRI image stimulation dataset, is applied in the experiment. During the dataset creation process, subjects in each session are guided to observe a sequence of images, and are asked whether the current image has shown before. The fMRI signals during the observation of each image are recorded. As illustrated in Figure \ref{fig:ana}(a), by using the templates provided by NSD, the 3D fMRI data collected from one specific subject can be converted into vectors, yielding sequences $F_1,F_2,\dots,F_n$. These vectors are regraded as input to a pre-trained CLIP image encoder for obtaining meaningful embeddings of each image. The corresponding CLIP embedding sequence is written as $C_1,C_2,\ldots,C_n$, which will be used in the subsequent analyses.

\paragraph{Ridge Regression Analysis}
To explore the amount of past information retained in brain signals, we formulate it as analyzing the correlation between the fMRI vector $F_t$ at the current time step $t$ and the CLIP embeddings $C_{t-k}$ of different past time points. The offset $k$ measures the number of time steps. For example, $k=0$ means the current fMRI vector is paired with the current CLIP embedding, and $k=1$ means the current fMRI vector is paired with last time point's CLIP embedding. As the time span increases, offset $k$ changes from 0 to $max_k$. 
Ridge regression, a method to handle multicollinearity by introducing a regularization term to stabilize the model, is employed to investigate the correlation between $F_t$ and $C_{t-k}$. Ridge regression is performed for each $\left \langle F_t,C_{t-k} \right \rangle$ pair respectively as $k$ varies. The process is illustrated in Figure \ref{fig:ana}(b). 

We set $max_k=9$ and explore different $k$ values from 0 to $max_k$, where  Notably, since each image in the NSD is presented three times and appears randomly in the image sequence, we ensure that images in the test set are not included in the training set. Afterwards, we sequentially apply ridge regression analysis with $k$ ranging from 0 to $max_k$ to the partitioned data. Additionally, we establish a lowerbound by randomly matching all brain signals and CLIP embeddings (adhering to the test set division principle) and also performing ridge regression analysis. 

The results of test set for different $k$ values are shown in Figure \ref{fig:pcc}, with the x-axis representing different $k$ values, and the 'rand' representing our lowerbound result. From the results, it can be observed that as the value of $k$ increases, the correlation between brain signals and corresponding CLIP representations gradually decreases. Particularly at $k=3$, the correlation approaches the lower bound, indicating that the information contained in the brain signals related to more than three items is minimal or challenging to extract. 

\paragraph{Trial-wise Representational Similarity Analysis}
Another method we used for memory retention analysis is trial-wise representational similarity analysis. The computation process for each session is illustrated in Figure \ref{fig:ana}(c). For each session, we construct two representational dissimilarity matrice (RDM), $RDM_f$ and $RDM_c$, using the sequences $F_1,F_2,\ldots,F_n$ and $C_1,C_2,\ldots,C_n$. In an RDM, the rows and columns represent the vectors (fMRI vectors or CLIP embeddings) corresponding to the stimuli, and the cell values indicate the dissimilarity between vectors (1 - the Pearson correlation coefficient $\rho$). Thus, an RDM contains the dissimilarity levels between every pair of stimuli (both $F_i$ and $C_i$ sequences) and is a symmetric $n \times n$ matrix. The computation process for $RDF_f$ is as follow, and the calculation for $RDM_c$ follows the same procedure.
\begin{equation}
RDM_f = J - \begin{bmatrix}
    \rho(F_1, F_1)  & \cdots & \rho(F_1, F_n) \\
    \rho(F_2, F_1)  & \cdots & \rho(F_2, F_n) \\
    \vdots  & \ddots & \vdots \\
    \rho(F_n, F_1)  & \cdots & \rho(F_n, F_n)
\end{bmatrix}
\end{equation}

where \(J\) is an all-ones matrix.

RSA \cite{kriegeskorte2008representational} is well-suited for researchers to compare data across different modalities and even to bridge data from different species. Unlike traditional RSA based on the entire RDM matrix, our method focuses on the similarity representation between individual data trials. 
We calculate the trial-wise similarity between $RDM_f$ and $RDM_c$. Here, we also use the offset $k$, where the $t_{th}$ row of $RDM_c$ corresponds to the $(t+k)_{th}$ row of $RDM_f$. We compute their correlation coefficient $\rho_{k,t}$ and average the results of all rows that meet this requirement to obtain the trial-wise representational similarity score $\rho_{k,ave}$ for each session with offset $k$. Finally, we compute the average of values across all sessions to obtain the final trail-wise RSA score. 

The results are shown in Figure \ref{fig:rsa}. Similar to ridge regression analysis, trial-wise RSA also exhibits similar trends. Furthermore, we replace $RDM_c$ with $RDM_f$ to compute the similarity between current and past brain activity signals. We will further elaborate in Appendix \ref{sec:brsa}, with the results shown in Figure \ref{fig:brsa}.

\subsection{Task Description} \label{sec:task}
Based on the analysis above, we propose a task called ``Memory Disentangling.'' This task involves extracting information from past visual stimuli encoded in brain activity and separating this information from ongoing brain activity to mitigate the effects of proactive interference.

As shown in Figure \ref{fig:intro}, given the fMRI signal at time $t$, $F_t$, the task is to decode the image descriptions viewed at the current and previous total $(max_k+1)$ time points, denoted as Cap := \{$cap_t, cap_{t-1}, \ldots, cap_{t-max_k}$\}. Our analysis indicates that the brain signals at the current time primarily contain information about the last three moments, thus we set $max_k = 2$.

It is important to note that the Memory Disentangling task is not limited to decoding brain activity into descriptions of images, and it can also involve other forms of decoding such as image reconstruction. Unlike previous visual stimuli decoding tasks, the core challenge here is to capture information about multiple past moments from a single time point's fMRI signal and to remove the interfering parts of past information, thereby enabling higher quality information decoding. Additionally, due to the low temporal resolution of fMRI signals, brain activity between two scan frames may be lost. Memory Disentangling, which focuses on extracting past information, might help to supplement the missing information between scan frames. This could alleviate the issue of fMRI's temporal resolution and enhance the development of brain decoding under continuous stimulation.

\section{Method} 
\begin{figure}[t]
  \includegraphics[width=\columnwidth]{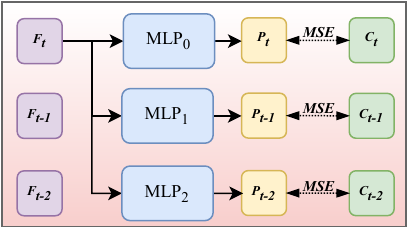}
  \caption{A schematic diagram of the straightforward approach using separate MLPs.}
  \label{fig:stra}
  \vspace{-1em}
\end{figure}
\begin{figure*}[t]
\centering
  \includegraphics[scale=1.05]{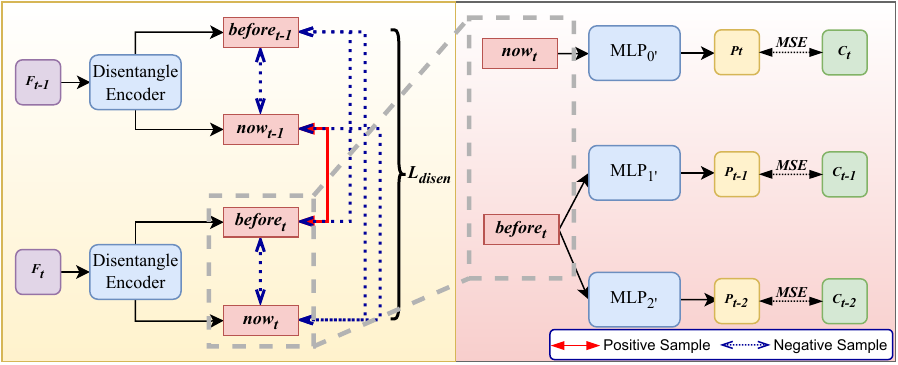} 
  \caption {Overview of the disentangled contrastive learning method. The left half illustrates the example of disentangled contrastive learning, while the right half shows the mapping process to the target image CLIP representations. Further explanation of the division of positive and negative sample pairs in disentangled contrastive learning is provided in Appendix \ref{sec:add}.}
    \label{fig:dis}
    \vspace{-1em}
\end{figure*}
In this section, we propose a method to address Memory Disentangling task, starting with a straightforward method. Initially, we employ separate Multilayer Perceptrons (MLPs) for predicting embeddings corresponding to each moment based on the current fMRI signal. This approach is easy to implement but has significant limitations as it does not leverage the memory relational memory dynamics between successive fMRI signals. To overcome these limitations, we design a disentangled contrastive learning method based on the theory of proactive interference.

\subsection{Straightforward Method Using Separate MLPs} \label{sec:stra}
For this task, we reformulate it as predicting the CLIP embeddings \( C_{t-k} \) for the current and the preceding $k_{th}$ moment using the current fMRI signal \( F_t \), followed by generating image captions using a pre-trained CLIP-to-caption model. The schematic diagram of the straightforward method is displayed in the Figure \ref{fig:stra}. Thus, the key point of the task lies in mapping the fMRI signal \( F_t \) to the CLIP embeddings \( C_t, C_{t-1}, C_{t-2} \). This method is to assign an MLP, denoted as \( MLP_k \), for the mapping of the \( k_{th} \) past moment. By inputting \( F_t \) into each \( MLP_k \), we obtain $k$ outputs, where each is subjected to an MSE loss with its corresponding \( C_i \), and the losses are summed up for training. The formula for the loss is as follows:
\begin{equation}
\mathcal{L}_{mse} = \sum_{i=0}^{max_k} MSE_i(P_{t-i}, C_{t-i})
\end{equation}
After that, for each CLIP embedding, the Pre-trained CLIP-to-caption model can generate corresponding image caption.
\subsection{Disentangled Contrastive Learning}
Motivated by the desire to disentangle the information from past memories embedded in the fMRI signal \( F_t \), we propose a disentangled contrastive learning approach based on memory proactive interference theory, which posits that cognitive processes are subject to the influence of previously acquired knowledge. Its core idea is that the memory component of current brain signals closely resembles the stimuli seen in the previous moment, a relationship present in all adjacent fMRI signals, thereby exhibiting continuity. That is, the neural representation of past information at time \textbf{t} is hypothesized to bear a closer resemblance to the current information at time \textbf{t-1}. 


Accordingly, we introduce a contrastive learning method to disentangle the brain signal into ``before'' and ``now'' components of semantic information, which we term ``disentangled contrastive learning.'' This enables the fMRI disentangle encoder to learn to disentangle past components. Subsequently, we use MLPs for mapping as in the first part, and this process is depicted in Figure \ref{fig:dis}.

For the disentangled contrastive learning, we input consecutive fMRI signals \( F_{t-1}, F_t \) into the same fMRI disentangle encoder, yielding four components: \( before_{t-1}, now_{t-1}, before_{t}, \) and \( now_{t} \). We set \( now_{t-1} \) and \( before_{t} \) as positive samples, with all other pairings as negative samples, and then employ an InfoNCE \cite{oord2018representation} loss for training. For simplicity, we denote \(\text{before}_{t-1}\) as \(b_{t-1}\), \(\text{before}_{t}\) as \(b_{t}\), \(\text{now}_{t-1}\) as \(n_{t-1}\), and \(\text{now}_{t}\) as \(n_{t}\).

To compute the similarity for positive pairs, we first calculate the cosine similarity between \(b_t\) and \(n_{t-1}\), denoted as \(s(b_t, n_{t-1})\):
\begin{equation}
s(b_t, n_{t-1}) = \frac{b_t \cdot n_{t-1}}{\|b_t\| \|n_{t-1}\|}
\end{equation}
where \(b_t \cdot n_{t-1}\) represents the dot product of \(b_t\) and \(n_{t-1}\), and \(\|b_t\| \) and \(\|n_{t-1}\|\) denote the norms of \(b_t\) and \(n_{t-1}\), respectively.

To clarify the negative samples formed by two consecutive moments, we refer to Table \ref{tab:neg}. We also calculate their similarities in the same manner.

\begin{table}[h]
\centering
\begin{tabular}{|c|c|}
\hline
Negative Pair 1 & $(n_t, b_t)$ \\ \hline
Negative Pair 2 & $(n_{t-1}, b_{t-1})$ \\ \hline
Negative Pair 3 & $(n_t, n_{t-1})$ \\ \hline
Negative Pair 4 & $(b_t, b_{t-1})$ \\ \hline
Negative Pair 5 & $(n_t, b_{t-1})$ \\ \hline
\end{tabular}
\caption{Negative pairs for disentangled contrastive learning} 
\vspace{-2em}
\label{tab:neg}
\end{table}
Finally,the InfoNCE loss is defined as:
\begin{equation}
\begin{split}
\mathcal{L}_{\textit{InfoNCE}} =
- \log \frac{\exp\left(s(b_{t}, n_{t-1}) / \tau\right)}{\sum_{(x, y) \in S} \exp\left(s(x, y) / \tau\right)} 
\end{split}
\end{equation}
where $\tau$ is a temperature parameter that controls the concentration of the distribution.

The final training loss $\mathcal{L}$ is a combination of the MSE loss and the InfoNCE loss:
\vspace{-0.5em}
\begin{equation}
\mathcal{L} = \mathcal{L}_{mse} + \alpha \mathcal{L}_{InfoNCE}
\vspace{-0.5em}
\end{equation}
where $\alpha$ is a weighting factor for $\mathcal{L}_{InfoNCE}$.

\subsection{Semantic Feature Decoding}
The two methods described above convert fMRI signals into CLIP embeddings, representing the semantic information of the visual stimuli at various time points. Subsequently, we need to transform these embeddings into textual descriptions, which are easier to observe and evaluate. CLIPCap \cite{mokady2021clipcap} is an image captioning model that generates descriptions from the CLIP embeddings of images. Given its superior performance, we use a pre-trained CLIPCap model to generate descriptions from our predicted CLIP embeddings. Consequently, we can ultimately convert the current fMRI signals into textual descriptions of the visual stimuli from the past few moments.

\begin{figure*}[t]
\centering
  \includegraphics[scale=0.42]{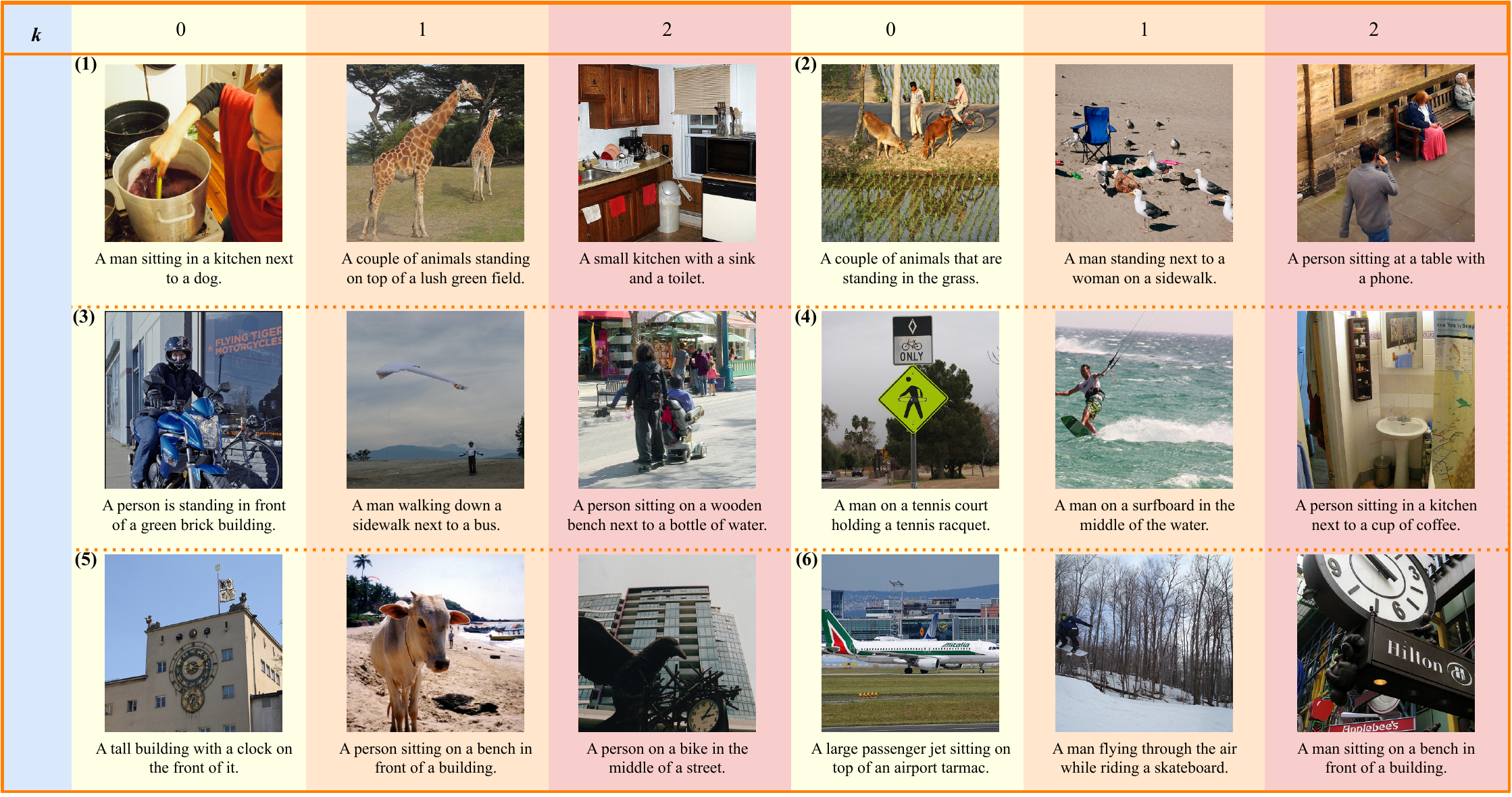} 
  \caption {Qualitative results of Memory Disentangling for Subject 1. Each example includes the decoded results at three different time points along with their corresponding visual stimuli.}
    \label{fig:result}
\end{figure*}

\begin{figure}[t]
  \includegraphics[width=\columnwidth]{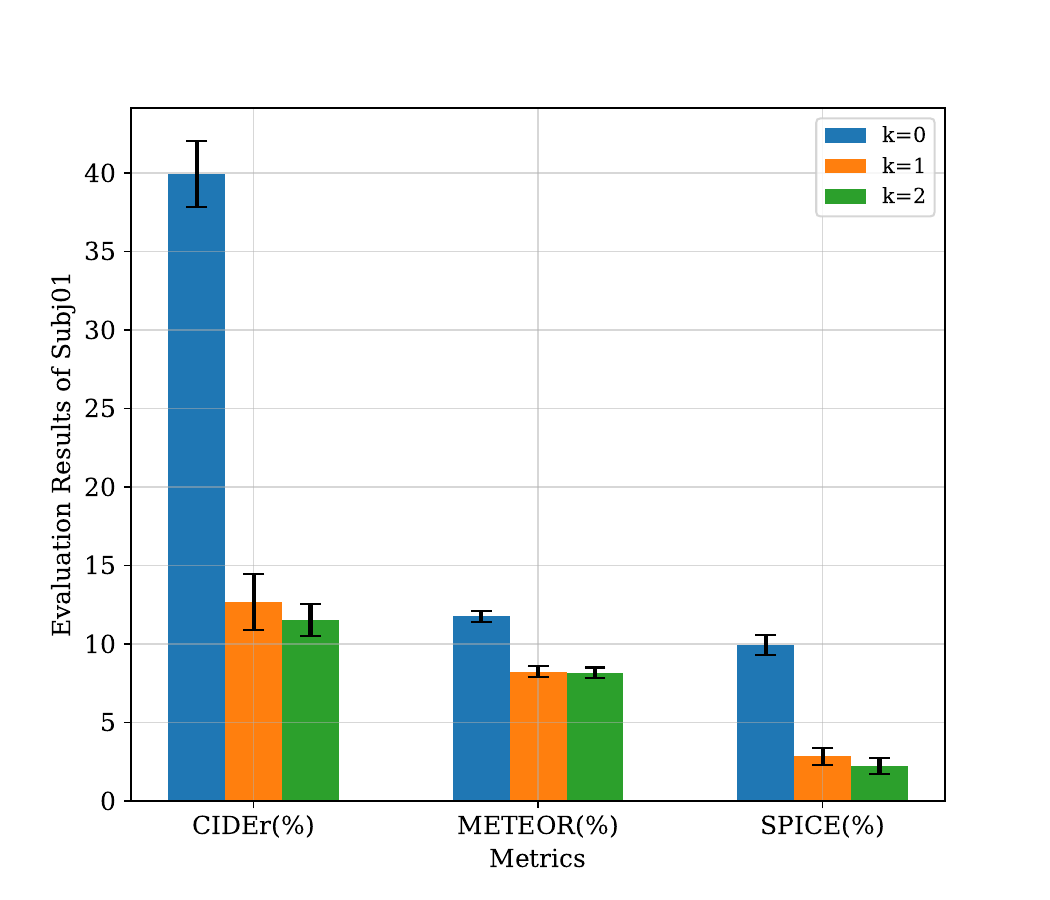}
  \caption{Quantitative results of Memory Disentangling for Subject 1.}
  \vspace{-1em}
  \label{fig:eval}
\end{figure}

\section{Experimental Settings}
\subsection{Dataset and Processing}
In our study, we utilize the Natural Scenes Dataset (NSD), a large-scale dataset of fMRI scans in response to visual stimuli from MS COCO dataset \cite{lin2014microsoft}. This dataset includes fMRI scans from eight subjects, obtained using a 7-Tesla fMRI scanner. During the scans, subjects are asked to view images and judge whether they have seen the presented image before. Each subject observes 9,000-10,000 distinct images, with each image appearing three times, randomly distributed throughout the image sequence. In the experiment, each subject undergoes 30-40 scan sessions, with each session containing 12 scan runs. There is a rest period between each pair of runs, and each run contains a continuous sequence of 62-63 images. Therefore, each session contains a total of 750 images. During each scan, image is presented for 3 seconds, followed by a 1-second blank screen. For more detailed information about the dataset, please visit the official website\footnote{\url{http://naturalscenesdataset.org}}.

In our study, we use the data from subject 1, 2, 5, and 7, as they have complete image scanning sessions, totaling 27,750 trials. We utilize the preprocessed functional scans at a resolution of 1.8 mm provided by NSD, along with the predefined template \texttt{nsdgeneral} to obtain fMRI vectors.

For each fMRI signal in the session, we use a sliding window of size 3 to store the CLIP image embeddings of continuous visual stimuli. Additionally, to ensure the images in the window are temporally contiguous, any data where the images span two different runs (indicating a long interval between stimuli presentations) are removed. Since each image is presented three times, it is crucial to strictly control data splitting to prevent contamination.  We first select a test dataset of size $m$ by randomly choosing $m$ data pairs, each pair containing the fMRI signal $F_t$ and the CLIP representations of images at times $t-2, t-1$, and $t$, in the form of $\left \langle F_t;C_{t},C_{t-1},C_{t-2} \right \rangle$. Images appearing in the test set are marked. Subsequently, we evaluate the remaining data, discarding any data points where the images in the window are already marked, thereby obtaining the training set. This process ensures that the training data is free from contamination.

\setlength{\tabcolsep}{1.5pt}

\begin{table*}[!t]
\centering
\small
\begin{tabular}{clccc|ccc|ccc}
\toprule
Metrics && \multicolumn{3}{c}{CIDEr(\%)} & \multicolumn{3}{c}{METEOR(\%)} & \multicolumn{3}{c}{SPICE(\%)} \\  \cmidrule(r){1-5}\cmidrule(r){6-8}\cmidrule(r){9-11}
\multicolumn{1}{c|}{} & \multicolumn{1}{c|}{\diagbox[width=3em,height=2em]{$\alpha$}{$k$}} & 0 & 1& 2  & 0 & 1 & 2 & 0 & 1 & 2       \\ \midrule

\multicolumn{1}{c|}{SF}& \multicolumn{1}{c|}{\_} &$34.3_{\pm 3.16}$ &$12.9_{\pm 2.25}$&$11.0_{\pm 0.99}$&$11.3_{\pm 0.41}$&\bm{$8.52_{\pm 0.28}$}&$8.24_{\pm 0.17}$& $8.68_{\pm 1.06}$&$2.85_{\pm 0.92}$ & $1.91_{\pm 0.45}$ \\\cmidrule(r){1-2}\cmidrule(r){3-5}\cmidrule(r){6-8}\cmidrule(r){9-11}

\multicolumn{1}{c|}{Ours} & \multicolumn{1}{c|}{0} &$35.1_{\pm 4.46}$&\bm{$13.4_{\pm 0.70}$}&\bm{$11.7_{\pm 0.40}$}&$11.3_{\pm 0.55}$&$8.14_{\pm 0.42}$&$8.23_{\pm 0.11}$&$9.31_{\pm 1.09}$&\bm{$3.21_{\pm 0.40}$}&\bm{$2.29_{\pm 0.36}$ }\\\cmidrule(r){1-2}\cmidrule(r){3-5}\cmidrule(r){6-8}\cmidrule(r){9-11}

\multicolumn{1}{c|}{Ours}& \multicolumn{1}{c|}{0.01} &\bm{$39.9_{\pm 2.11}$}&$12.6_{\pm 1.76}$&$11.5_{\pm 1.0}$&\bm{$11.7_{\pm 0.35}$}&$8.24_{\pm 0.33}$&$8.15_{\pm 0.32}$&\bm{$9.95_{\pm 0.64}$}&$2.84_{\pm 0.52}$&$2.25_{\pm 0.51}$ \\\cmidrule(r){1-2}\cmidrule(r){3-5}\cmidrule(r){6-8}\cmidrule(r){9-11}

\multicolumn{1}{c|}{Ours}& \multicolumn{1}{c|}{0.1}  & $38.5_{\pm 6.91 }$&$10.9_{\pm 0.99}$&$11.7_{\pm 0.94}$&$11.4_{\pm 0.73}$&$8.18_{\pm 0.2}$&\bm{$8.27_{\pm 0.29}$}&$9.45_{\pm 1.11}$&$1.80_{\pm 0.53}$  & $1.83_{\pm 0.60}$\\\bottomrule 
\end{tabular}
\caption{Ablation study results, including the straightforward method and our proposed disentangled contrastive learning method. The symbol $\alpha$ represents the weight of the loss $\mathcal{L}_{InfoNCE}$. 'SF' stands for Straightforward method mentioned in Section \ref{sec:stra}, and the best result for each $k$ is bolded.}
\vspace{-1em}
\label{tab:abla}
\end{table*}

\subsection{Implementation}
In our task analysis section, we employ the NeuroRA \cite{lu2020neurora} toolkit to compute the RDM matrix. Regarding Memory Disentangling, we opt for a $L_{infonce}$ weight $\alpha$ of 0.01, a selection derive from a ablation study in Section \ref{sec:abla}. The size of our testing data $m$ is set to 500, and a validation set of the same size was randomly selected from the partitioned training set. During the training phase, we optimized the model using AdamW \cite{loshchilov2019decoupled} with an initial learning rate of 1e-5. We employ 5 different seeds for partitioning and training to enhance the reliability of our results. The reported experimental outcomes represent the average of these results obtained from 5 random seeds.
\subsection{Evaluation Metrics}
Since we employ the pre-trained CLIPCap model to generate image captions from the predicted disentangled outputs, semantics-based evaluation becomes more appropriate. To evaluate the degree of matching between the generated captions and the images, we obtain the COCO captions for all images in test set, and use the CIDEr \cite{vedantam2015cider}, METEOR \cite{denkowski2014meteor}, and SPICE \cite{anderson2016spice} metrics for evaluation.  For the calculation of evaluation metrics, we use the nlg-eval\footnote{\url{https://github.com/Maluuba/nlg-eval}} library, which is specifically designed for NLG evaluation.

\section{Results and Analysis}

\subsection{Qualitative Results}

To provide an intuitive understanding of the Memory Disentangling task under continuous visual stimuli, Figure \ref{fig:result} reports several decoding examples from Subject 1. Each example includes semantic decoding results at three time points, representing the caption results decoded at the current, and two previous moments. 

The results indicate that decoding at the current moment ($k=0$) yields partially accurate results. However, the accuracy for past moments is relatively poor, with a tendency to set the subject as ``\texttt{a person/man ...}''.  This may be due to the high frequency of human figures in the dataset. Additionally, the decoding results tend to be broad; for example, at time point 1 in the first sample and time point 0 in the second sample, giraffes and cows are both decoded as ``animal''. Occasionally, descriptions matching the images are obtained, indicating that decoding past information remains challenging. Overall, the decoding results exhibit discrepancies with the visual stimuli, attributed to the low signal-to-noise ratio (SNR) of brain activity signals. This low SNR makes brain decoding highly challenging, and extracting past information from these signals even more difficult.

\subsection{Quantitative Results}
Figure \ref{fig:eval} presents the evaluation of decoding results of Subject 1 at three time points across different metrics, including CIDEr, METEOR, and SPICE. Consistent with intuition, all metrics show varying degrees of decline over time (represented in the figure as increasing $k$ values), with the CIDEr metric showing the most pronounced drop. This trend suggests that the accuracy and richness of the decoded descriptions deteriorate as the temporal distance from the current moment increases. Additional quantitative results for other subjects are provided in Appendix \ref{sec:other}.

\subsection{Ablation Study} \label{sec:abla}
We conducted multiple experiments on the NSD dataset to perform Memory Disentangling task. We evaluated both the straightforward method and our proposed disentangled contrastive learning method with varying loss weight schemes represented by $\alpha$. The experimental outcomes are summarized in Table \ref{tab:abla}.

The disentangled contrastive learning method, with a loss weight of 0.01, consistently achieves optimal results at the current time point, demonstrating its positive role in removing past interference from fMRI brain signals. This effect may mitigate some of the effects of proactive interference on memory. However, this was not reflected in the decoding at the subsequent two time points, indicating a need for further improvement in extracting past information, and a relatively large weight of $\mathcal{L}_{InfoNCE}$ might impede the mapping from fMRI to CLIP space, resulting in reduced decoding performance and increased variance. We speculate that this might be due to some information loss in the representation of extracted past information.

\section{Conclusion}
This study proposes the task of Memory Disentangling by analyzing the past information contained in working memory under continuous visual stimuli. Based on the phenomenon of proactive interference, we introduce a disentangled contrastive learning method to complete the Memory Disentangling task, which involves decoding semantic content at multiple time points from fMRI signals and remove the interfering parts of past information. This approach may help alleviate the low temporal resolution of fMRI and contribute new insights to the field of brain decoding.

\section*{Limitations}
Although we explored the content of past information contained in fMRI signals, its interpretability remains limited. Additionally, while we focused on semantic decoding for Memory Disentangling tasks, we did not address other forms of memory disentanglement, such as image reconstruction. While our proposed disentangled contrastive learning method showed improvement in current-time decoding, its effectiveness in extracting past memory information was suboptimal, necessitating further in-depth exploration in future research. Specifically, there is a need to investigate how to optimize models to better capture past memory accurately and to enhance the model's ability to learn from brain signals. Furthermore, expanding to other Memory Disentangling tasks would help comprehensively assess the method's generality and applicability, thus advancing the field of cognitive neuroscience.

\section*{Ethical Statement}
We are committed to maintaining the highest standards of ethical conduct in our research endeavors. The dataset used in this study adheres strictly to ethical guidelines, encompassing rigorous informed consent protocols, participant confidentiality safeguards, and meticulous data handling practices. Our commitment to transparency ensures that all research procedures are conducted with integrity, while prioritizing the security and privacy of participant data. By adhering to these ethical standards, we aim to responsibly advance the fields of neuroscience and artificial intelligence, contributing meaningfully to scientific knowledge and societal well-being.
\section*{Acknowledgements}
This research is supported by the National Natural Science Foundation of China (No.62476127, No.62106105),  the Natural Science Foundation of Jiangsu Province (No.BK20242039), the CCF-Baidu Open Fund (No.CCF-Baidu202307), the CCF-Zhipu AI Large Model Fund (No.CCF-Zhipu202315), the Fundamental Research Funds for the Central Universities (No.NJ2023032), the Scientific Research Starting Foundation of Nanjing University of Aeronautics and Astronautics (No.YQR21022), and the High Performance Computing Platform of Nanjing University of Aeronautics and Astronautics.
\bibliography{custom}

\appendix

\begin{figure*}[t]
\centering
  \includegraphics[scale=1.2]{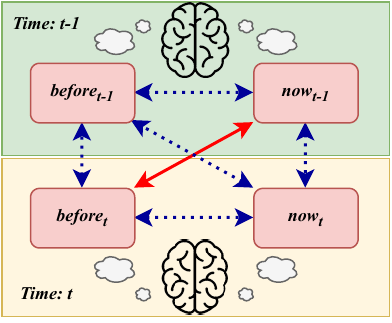}
  \caption{Further explanation of the positive and negative sample pairs division in Figure \ref{fig:dis}. The positive and negative sample pairs are represented by red solid lines and blue dashed lines, respectively, consistent with Figure \ref{fig:dis}.}
  \label{fig:add}
\end{figure*}
\section{Related Work} \label{sec:rela}
\subsection{Brain Decoding of Visual Stimuli}
Brain decoding aims to interpret neural activity patterns and link them to perceptual, cognitive, or motor processes. Recent advancements in neuroimaging technologies, particularly functional magnetic resonance imaging (fMRI), significantly enhance our ability to decode visual stimuli. \cite{haxby2001distributed} make groundbreaking progress in this field by revealing how the ventral temporal cortex encodes different categories of visual objects. \cite{kamitani2005decoding} use multivoxel pattern analysis (MVPA) to classify the direction of visual stimuli based on brain activity. Building on this early work, researchers explore various brain decoding tasks, such as stimulus classification and reconstruction.

Stimulus classification involves categorizing different types of visual stimuli based on brain activity. \cite{yargholi2016brain} utilize an enhanced Naive Bayes classifier to decode handwritten digits from fMRI data. The current focus in brain decoding shifts more towards image reconstruction. Early image reconstruction techniques use linear regression models to map fMRI signals to given image features \cite{naselaris2009bayesian,kay2008identifying}. With the advancement of deep learning technologies, more image reconstruction methods now employ Latent Diffusion Models (LDM) with image generation capabilities, achieving high-quality reconstruction results \cite{scotti2024reconstructing, ozcelik2023natural, sun2024contrast}. Additionally, describing the content of images seen by subjects from brain signals can be viewed as a form of reconstruction—semantic reconstruction of images. \cite{chen2023mindgpt} use cross-attention and GPT-2 to accomplish semantic reconstruction tasks.

Besides the reconstruction of static visual stimuli, some researchers also tackle the reconstruction of continuous visual stimuli. The low temporal resolution of fMRI makes this task particularly challenging. \cite{chen2024cinematic} use contrastive learning to map fMRI to the CLIP representation space, fine-tuning Stable Diffusion on a video-text dataset to successfully reconstruct coherent videos with clear semantic information.
\subsection{Tracking Visual Memory through Brain Activity Patterns}
Research on visual memory trajectories focuses on decoding and tracking the storage and recall processes of visual stimuli in memory through brain activity patterns. This area of study not only enhances our understanding of how visual information is encoded and stored in the brain but also reveals the dynamic changes during memory retrieval. \cite{davis2021visual} employ fMRI and item-wise RSA to investigate how memory representations generated during the encoding of individual items influence subsequent contextual memory. \cite{luo2023representational} utilize electroencephalography (EEG) recordings and RSA analysis to explore the neural basis of sequential dependency in visual perception. Their findings indicate that EEG signals retain information about previously seen objects, which affects current perceptual responses. \cite{fafrowicz2023dynamics} use fMRI and Independent Component Analysis (ICA) to study the formation of working memory and false memory. Additionally, time-of-day effects are observed, influencing memory-related brain network activity and performance.
\begin{figure}[t]
  \includegraphics[width=\columnwidth]{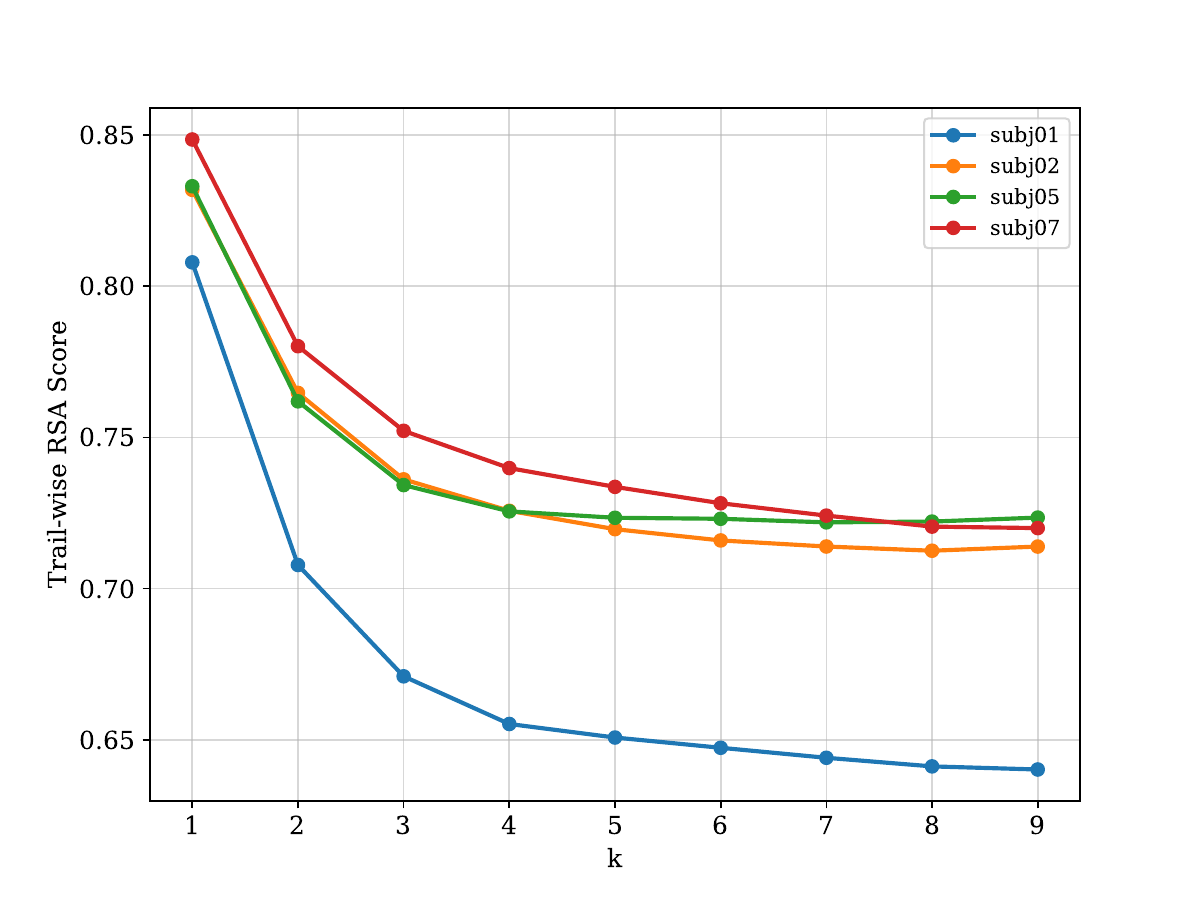}
  \caption{Trial-wise RSA Results of Current versus Past Brain Activity.}
  \label{fig:brsa}
\end{figure}
\section{Similarity Analysis of Current and Past Brain Activity Signals} \label{sec:brsa}
Since fMRI indirectly measures neural activity in the brain by detecting BOLD signals, and changes in blood oxygen levels and blood flow occur gradually and continuously, fMRI data also exhibit a certain level of continuity. We alse use trial-wise RSA to explore the correlation between brain activities at different times. Specifically, we replace $RDM_c$ in Figure \ref{fig:ana}(c) with $RDM_f$, starting from $k=1$ (since $k=0$ represents the correlation between the current time and the current brain signal, which is always 1), keeping the rest of the operations unchanged. The RSA results between current and past brain activities are shown in Figure \ref{fig:brsa}.

From the figure, it can be observed that there is a gradual decrease in correlation between consecutive fMRI signals, and the RSA scores for each participant begin to stabilize around $k=4$.
\section{Results from Other Subjects} 
\label{sec:other}
\begin{figure}[t]
  \includegraphics[width=\columnwidth]{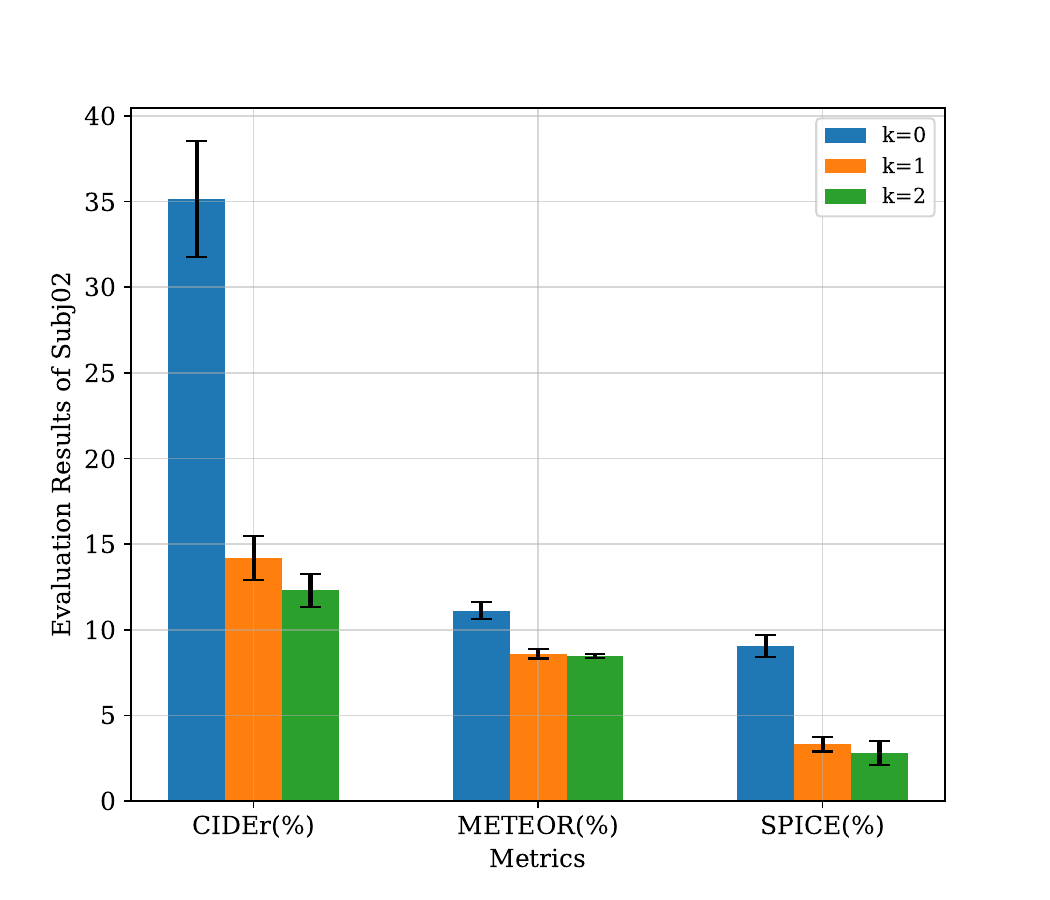}
  \caption{Quantitative results of Memory Disentangling for subject 2.}
  \label{fig:eval02}
\end{figure}

\begin{figure}[t]
  \includegraphics[width=\columnwidth]{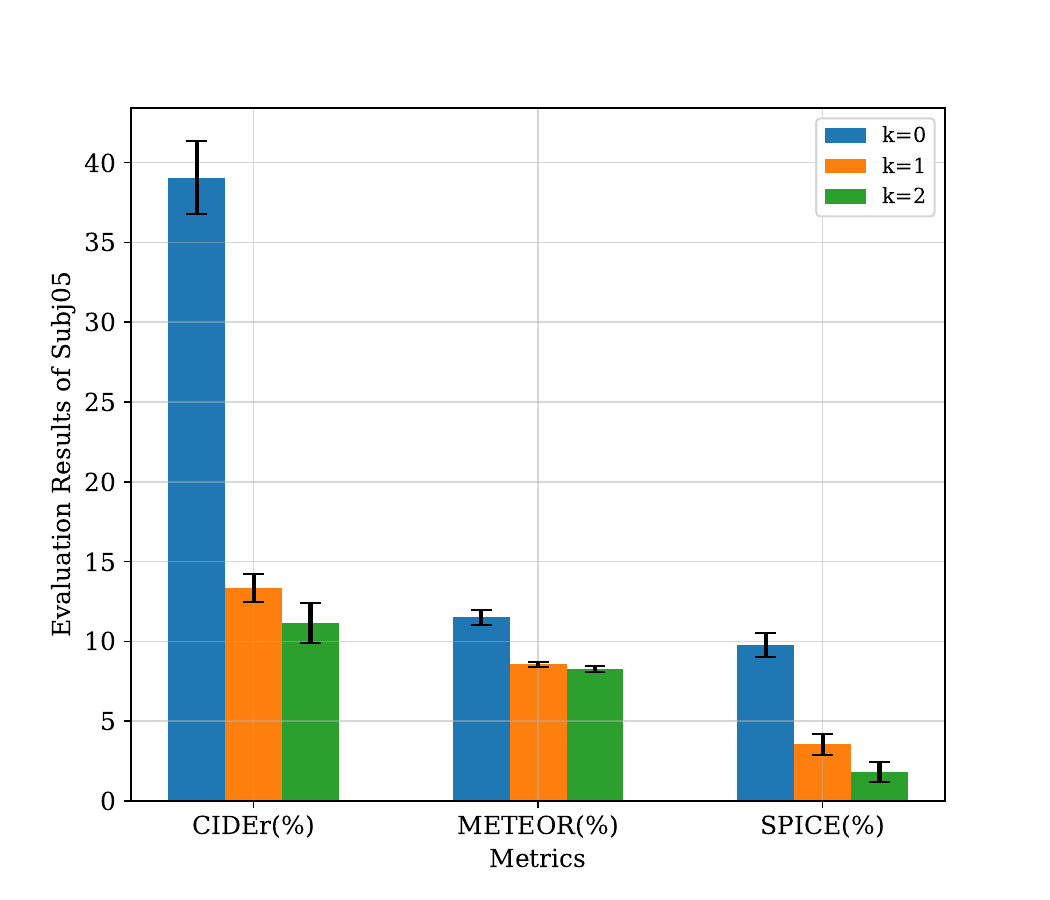}
  \caption{Quantitative results of Memory Disentangling for subject 5.}
  \label{fig:eval05}
\end{figure}

\begin{figure}[t]
  \includegraphics[width=\columnwidth]{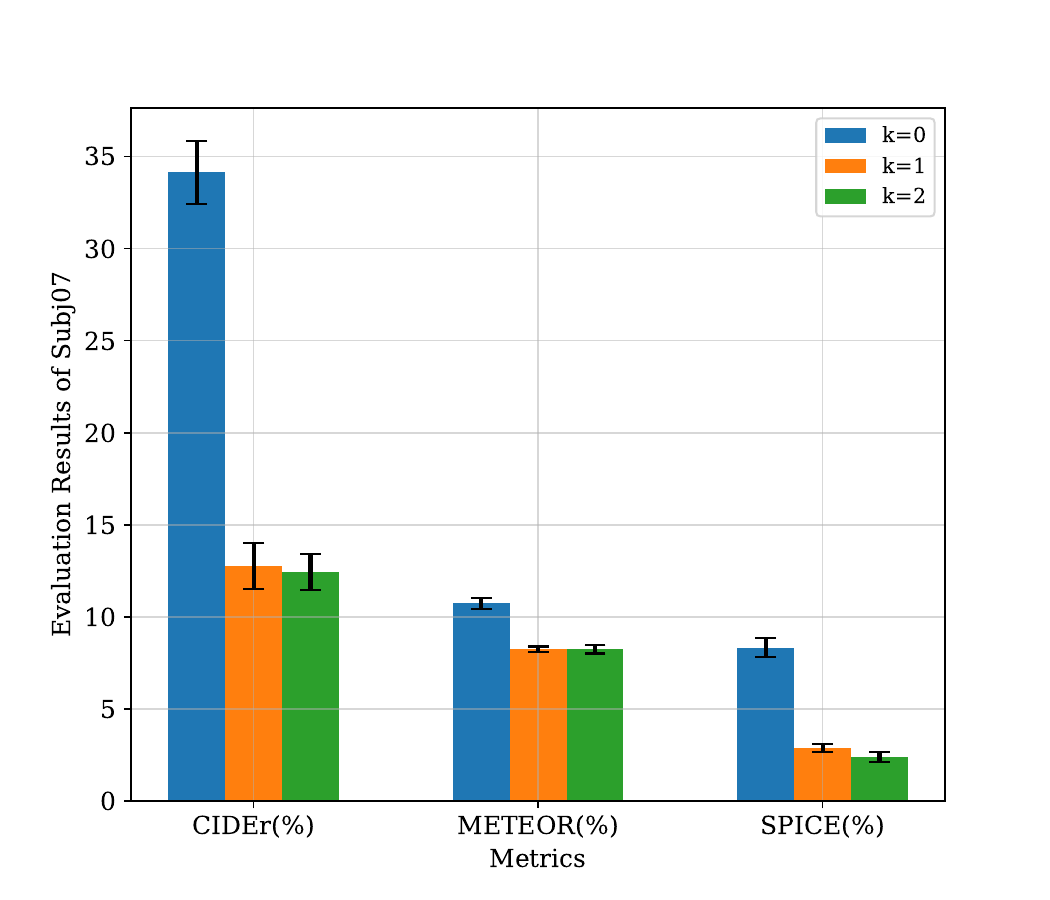}
  \caption{Quantitative results of Memory Disentangling for subject 7.}
  \label{fig:eval07}
\end{figure}

In this section, we present the quantitative decoding results for the remaining participants, depicted in Figure \ref{fig:eval02}, \ref{fig:eval05}, and \ref{fig:eval07}.
\section{Additional Explanation of Disentangled Contrastive Loss} \label{sec:add}
Due to the complexity of the positive and negative sample pairs division, in order to provide a clearer explanation of this section, we separately illustrate the positive and negative samples in Figure \ref{fig:add} based on Figure \ref{fig:dis}.

\end{document}